\documentclass[runningheads]{llncs}
\usepackage[T1]{fontenc}
\usepackage{graphicx}
\usepackage{amsmath}

\begin{document}
\title{Unmasking Deep Fakes: Leveraging Deep Learning for Video Authenticity Detection}
\titlerunning{Deep Learning for Video Authenticity Detection}
%
%
\author{
Sadia Ruhama\inst{1} \and
Sabrina Tajnim Sithi\inst{1} \and
Mahmudul Hasan\inst{1}\orcidID{0009-0000-3160-0322} \and
Chowdhury Mohammad Mutamir Samit\inst{1} \and
Oindrila Saha\inst{1}
}
\authorrunning{S. Ruhama et al.}
%
\institute{
BRAC University, Dhaka, Bangladesh \\
\email{
sadia.ruhama@g.bracu.ac.bd,
sabrina.tajnim.sithi@g.bracu.ac.bd, \\ 
mahmudul.hasan5@g.bracu.ac.bd,
mohammad.mutamir.samit@g.bracu.ac.bd,\\
oindrila.saha@g.bracu.ac.bd}
}
\maketitle              
\begin{abstract}
Deepfake videos, produced through advanced artificial intelligence methods now-a-days, pose a new challenge to the truthfulness of the digital media. As Deepfake becomes more convincing day by day, detecting them requires advanced methods capable of identifying subtle inconsistencies.  The primary motivation of this paper is to recognize deepfake videos using deep learning techniques, specifically by using convolutional neural networks. Deep learning excels in pattern recognition, hence, makes it an ideal approach for detecting the intricate manipulations in deepfakes.  In this paper, we consider using MTCNN as a face detector and EfficientNet-B5 as encoder model to predict if a video is deepfake or not. We utilize training and evaluation dataset from Kaggle DFDC. The results shows that our deepfake detection model acquired 42.78\% log loss, 93.80\% AUC and 86.82\% F1 score on kaggle's DFDC dataset. 

\end{abstract}

\section{Introduction}
The swift proliferation of smartphones and social media has resulted in an exponential growth in the creation and sharing of digital images and videos. Alongside this, advancements in artificial intelligence and computer vision have made sophisticated image and video maneuver methods more popular than ever before \cite{durall2019unmasking}. In the past, such modifications required extensive expertise, but with the advent of deep learning-based tools, even individuals with minimal technical knowledge can now alter digital content effortlessly \cite{kingra2023emergence}. One such emerging technology is Deepfake, an artificial intelligence-based method that generates highly realistic but synthetic images, videos, and audio \cite{abbas2024unmasking} by manipulating facial expressions or swapping identities seamlessly \cite{arshed2023unmasking}.

While Deepfake technology has promising applications in entertainment, filmmaking, and gaming, it also raises suggestive moral and security issues \cite{das2023unmasking}. The capability to create convincingly fabricated media has fueled the spread of misinformation, political disinformation, and identity fraud. Malicious actors exploit Deepfake technology to generate deceptive content, leading to reputational damage, blackmail, and cyber harassment \cite{rana2022deepfake}. A notable example is the case of journalist Rana Ayyub, whose manipulated video went viral in 2018, causing severe personal and professional harm \cite{taeb2022comparison}. The increasing misuse of Deepfake technology poses a serious threat to digital media authenticity and national security, with reports indicating a 67\% rise in security breaches related to manipulated media.

Deepfake technology primarily relies on deep learning models, especially Generative Adversarial Networks (GANs), to synthesize hyper-realistic content that is often imperceptible from genuine media \cite{agarwal2020detecting}. Although these techniques enhance visual effects in various industries, their potential for misuse necessitates the development of robust detection frameworks. Recent studies suggest that nearly 50\% of the billions of media files uploaded daily on online platforms undergo some form of manipulation \cite{akhtar2024video}. To counteract this growing issue, researchers have explored various detection methodologies, including AI-driven forensic analysis and blockchain-based authentication mechanisms such as Smart Contracts and Hyperledger Fabric \cite{rashid2021blockchain}. However, despite these efforts, Deepfake detection remains a challenging and evolving field that requires continuous improvement.

This paper intent to enhance the accuracy of deepfake detection by integrating two powerful neural network models: Multi-task Cascaded Convolutional Networks (MTCNN) for facial identification and EfficientNet for image classification. By leveraging a large and high-quality dataset, we hypothesize that our approach will effectively distinguish between authentic and manipulated media, as indicated by a low binary cross-entropy value. Through further optimization and testing, we anticipate that our framework will significantly improve the reliability of Deepfake detection, offering a robust solution to mitigate the risks associated with this technology. The key contributions of this paper are as follows:
\begin{itemize}
    \item We proposed a MTCNN-EfficientNetB5 fusion model to detect deepfake videos where MTCNN was utilized for face detection and EfficientNetB5 for feature extraction and classification.
    \item We have tested our model on the DFDC dataset where the proposed model outperformed many state-of-the art models without extensive data processing and ensembling models.
\end{itemize}

The remainder of this paper is structured as follows: Section 2 reviews prior research on Deepfake detection methods. Section 3 describes the dataset used in this study. Section 4 outlines the proposed methodology, including data preprocessing steps, model architectures and training. Section 5 presents the performance evaluation metrics and results analysis. Section 6 concludes the paper.

\section{Related Work}
Despite the availability of various deepfake detection methods, identifying deepfakes remains a significant challenge. This is primarily due to the sophisticated nature of deep neural networks, which make it difficult to trace manipulation artifacts \cite{janutenas2023deep}. Additionally, celebrities are frequent targets of deepfake manipulations, including face-swapping and lip synchronization, which current detection techniques often fail to accurately detect. To overcome these obstacles, researchers have looked into strategies like dual-stream learning methods to improve detection capabilities and protect individuals from malicious deepfake applications \cite{chu2022protecting}.

The increasing accessibility of cloud computing has further exacerbated the deepfake problem, as it enables individuals with no prior knowledge of computer vision or deep learning to generate highly realistic fake videos. Consequently, numerous detection models have been proposed, including transfer learning, autoencoders, and hybrid models combining CNNs and RNNs \cite{suratkar2023deep}. One notable detection framework, DFN, utilizes MobileNet as its base architecture, incorporating separable convolution and max-pooling layers. DFN has demonstrated superior performance compared to Xception and EfficientNet, indicating its potential for enhancing deepfake detection \cite{bansal2023real}.

In addition to AI-driven methods, biometric-based forensic techniques have arrived as a auspicious solution for deepfake detection. These approaches leverage two key methodologies: Static Biometric Analysis, which employs traditional facial recognition techniques to analyze still images in videos, and Temporal Behavioral Analysis, which examines facial movements and expressions over time to detect inconsistencies. Such methods enhance detection accuracy by considering both spatial and temporal aspects of deepfake content \cite{agarwal2020detecting}.

Deng, Suo, and Li's study, "Deep Fake Video Detection Based on EfficientNet-V2 Network," proposes an innovative technique that integrates longitudinal data from video sequences with the EfficientNet-V2 architecture to improve detection performance \cite{deng2022deepfake}. While their approach shows promise in enhancing feature extraction and detection efficiency, it lacks comprehensive validation in real-world scenarios. 

Tan and Le's research, "EfficientNet: Rethinking Model Scaling for Convolutional Neural Networks," introduces a systematic method for scaling CNNs to achieve state-of-the-art performance while optimizing computational resources \cite{tan2019efficientnet}. Their EfficientNet architecture employs a compound scaling algorithm that uniformly adjusts model depth, width, and resolution. This method enables EfficientNet to outperform traditional architectures such as ResNet and Inception in image classification tasks while requiring fewer computational resources. However, a limitation of this approach is its reliance on predefined scaling values, which may not be universally optimal. Additionally, its primary application to image classification limits its effectiveness for tasks involving spatial analysis, such as object detection. 

Maryam Taeb and Hongmei Chi’s study, "Comparison of Deepfake Detection Techniques through Deep Learning," evaluates various deep learning architectures for deepfake detection, with an emphasis on robustness against fraud, reliability, and computational efficiency \cite{taeb2022comparison}. Their work highlights the advantages and limitations of different models, including CNNs and RNN-based hybrid approaches, across multiple datasets. While the study provides valuable insights into best practices for deepfake detection, it lacks extensive video-based analysis and dataset diversity, which limits its applicability to a broader range of deepfake manipulations. 

Overall, ongoing advancements in deepfake detection continue to refine existing methodologies, but challenges remain in terms of scalability, robustness, and real-world applicability. Future efforts should focus on developing more efficient and resilient frameworks capable of detecting deepfakes across diverse scenarios while minimizing computational overhead.

\section{Dataset}
A comprehensive, high-quality, and diverse dataset is essential for any deep learning application to mitigate issues such as overfitting and enhance model generalizability. In this study, the Deepfake Detection Challenge (DFDC) dataset, provided by Kaggle, was selected as the primary data source. Several factors motivated this choice, including its public availability, extensive documentation, and substantial volume of high-quality data.

The DFDC dataset is the largest publicly accessible face-swap video dataset to date, comprising over 100,000 video clips featuring 3,426 paid actors. These videos were generated using a diverse set of techniques, including various Deepfake approaches, Generative Adversarial Network (GAN)-based methods, and non-learned techniques. This diversity ensures that models trained on the dataset are exposed to outputs from multiple deepfake creation technologies, promoting robust detection capabilities across different manipulation methods.

To facilitate ease of access, the dataset is segmented into smaller downloadable files, as its overall size is substantial. The complete training set contains 119,154 ten-second video clips representing 486 unique subjects. Of these, approximately 100,000 clips feature synthetic content, accounting for 83.9\% of the dataset. This extensive collection supports the development and evaluation of deep learning models aimed at detecting synthetic media with improved accuracy and generalization.

\section{Methodology}
This paper aimed to determine whether a video is authentic or generated using deepfake technology. Since deep learning models require image inputs, the video data is first converted into individual frames before being processed by the system. Each frame is then preprocessed which helps improve the performance of the model. Key facial regions are extracted from the frames using MTCNN and face detection techniques to focus on the most relevant areas for deepfake identification and saved the processed data as PNG files. The processed image frames then passed into the EfficientNet model for detection and classification. For each model output, instead of simple averaging, a heuristic strategy was applied. The overall workflow of the proposed deepfake detection system is depicted in Figure 1.


\begin{figure}
\includegraphics[width=1.0\textwidth]{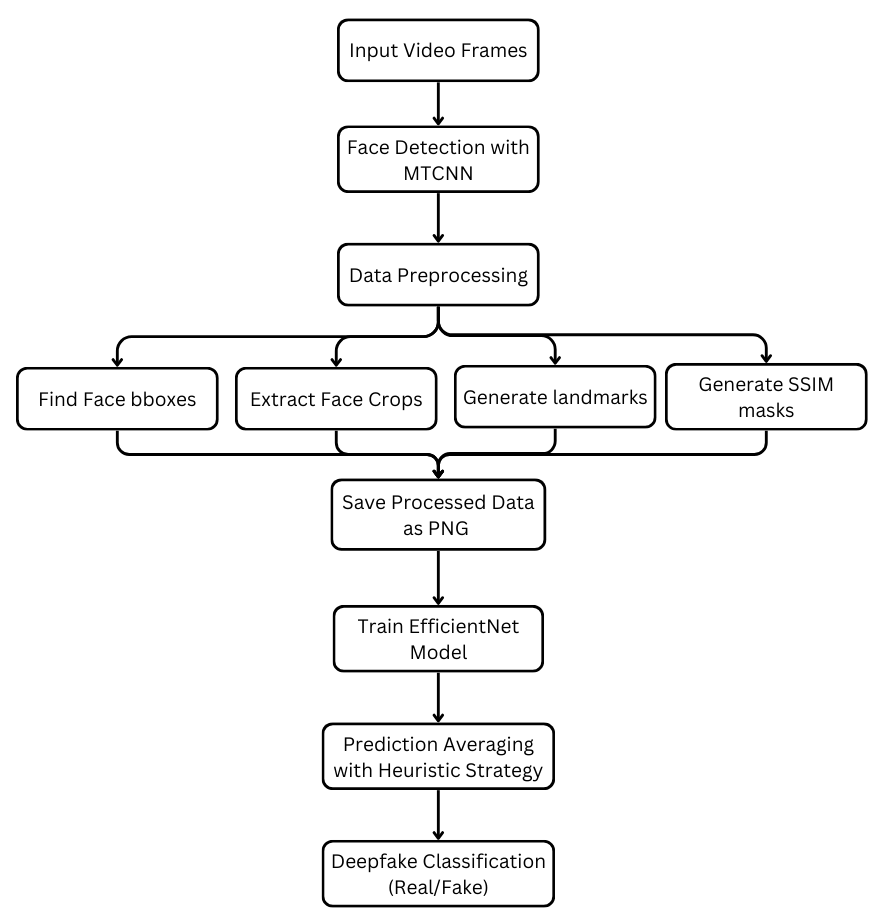}
\caption{Workflow Diagram} \label{fig1}
\end{figure}

\subsection{Data Preprocessing}
Data preparation and face detection were critical steps in this study, ensuring high-quality inputs for model training and evaluation. Initially, facial bounding boxes and landmarks were extracted from video frames using the MTCNN. The extracted boxes and landmarks were stored in JSON format for easy accessibility. Additionally, face crops were generated at their original resolution and saved as PNG images. To enhance the data, structural similarity (SSIM) masks were computed to highlight differences between real and fake frames, and these masks were also stored as PNG files. The MTCNN detector was employed for face detection due to its robustness and accuracy. Input size for the face detector was dynamically adjusted based on the video resolution to optimize performance and computational efficiency. The face detection steps are depicted in Figure 2 and the extracted face is shown in Figure 3.
\begin{figure}[htbp]
\centerline{\includegraphics[width=0.37\textwidth]{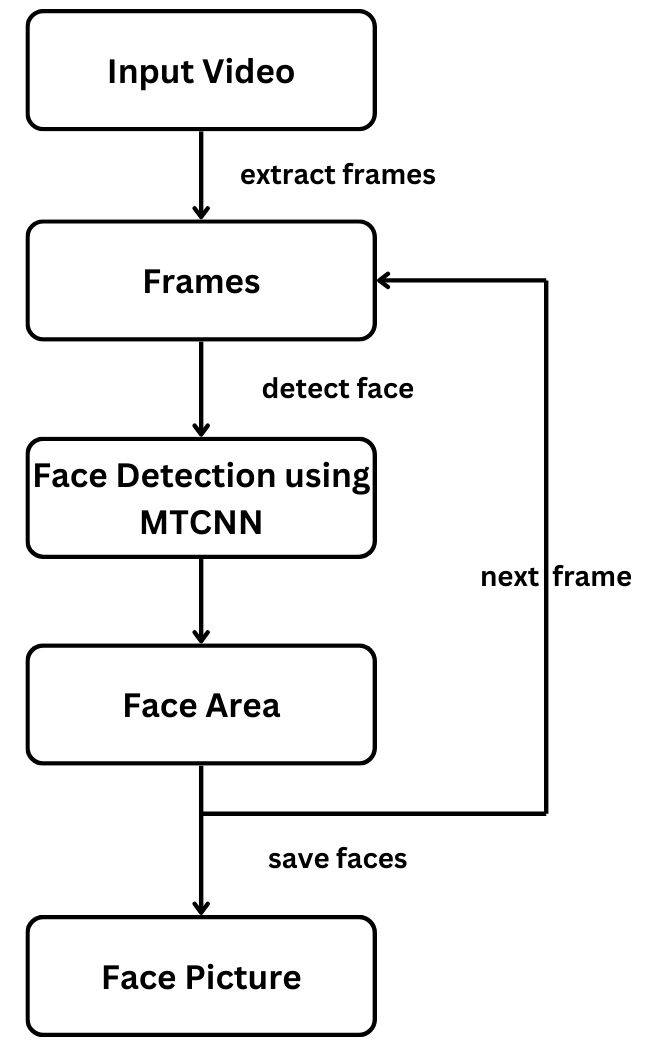}}
\caption{Face Detection Steps}
\label{fig2}
\end{figure}

\begin{figure}[htbp]
\centerline{\includegraphics[width=0.4\textwidth]{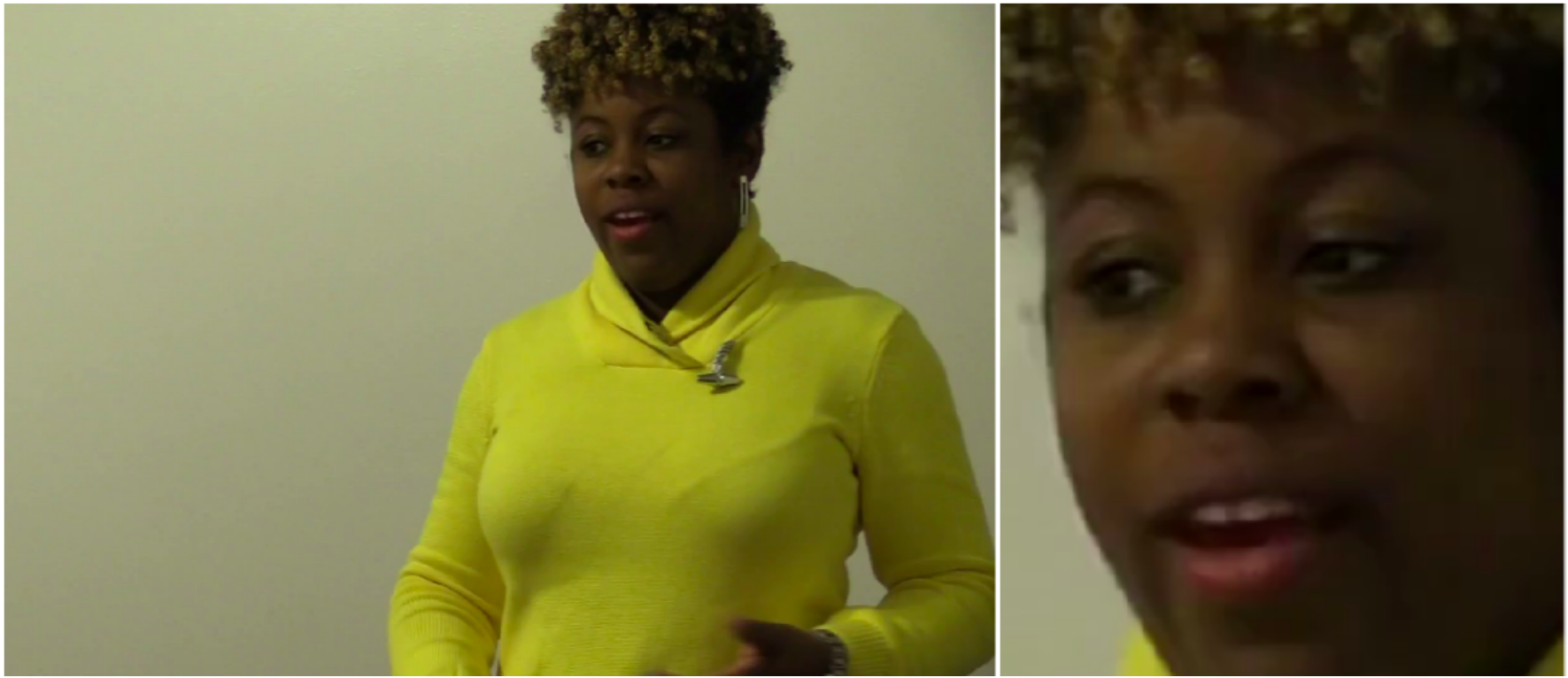}}
\caption{Extracted Frames (Left) \& Processed Images (Right)}
\label{fig3}
\end{figure}

Face detection and feature extraction are critical steps in the detection pipeline. MTCNN was selected as the face detection model, while EfficientNet served as the backbone for feature extraction and classification. The following sections describe these models and their roles in the system.

\subsection{MTCNN}
MTCNN is a widely-used face detection framework known for its accuracy and robustness. It detects faces under varying poses and lighting conditions while providing key facial landmarks essential for precise face alignment. MTCNN's efficiency in processing video frames with adjustable input sizes makes it suitable for handling diverse video resolutions. 

The MTCNN face detection process involves a three-stage cascaded network:
\begin{itemize}
\item P-NET: Generates candidate windows with bounding box regression vectors.
\item R-NET: Refines the candidate windows and filters non-face regions.
\item O-Net: Further refines face localization and identifies five facial landmarks.
\end{itemize}
This multi-stage process ensures accurate face extraction, enabling downstream analysis of facial regions for deepfake detection. The MTCNN architecture is illustrated in Figure 4.
\begin{figure}[htbp]
\centerline{\includegraphics[width=0.7\textwidth]{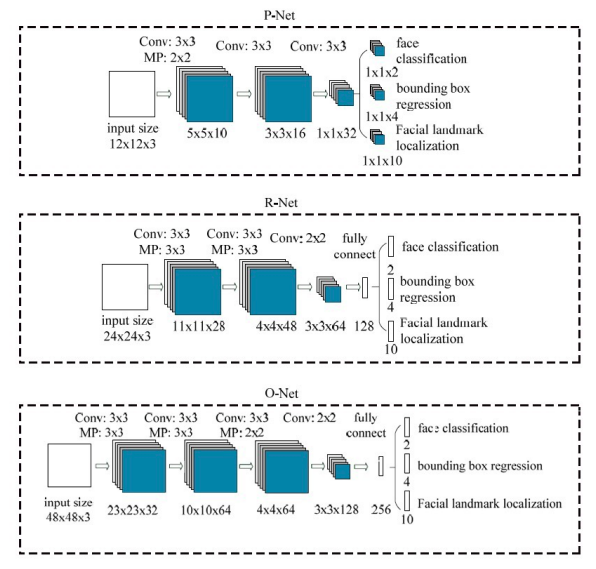}}
\caption{Architechture of MTCNN \cite{yuan2020minor}}
\label{fig4}
\end{figure}

\subsection{EfficientNet}
EfficientNet is a family of convolutional neural networks designed to achieve high accuracy with lower computational costs. Developed by Mingxing Tan and Quoc V. Le in 2019 \cite{tan2019efficientnet}, EfficientNet utilizes a compound scaling method to uniformly scale depth, width, and resolution based on a single coefficient. This approach results in models that are both compact and efficient compared to traditional architectures. The architechture of EfficinetNet is shown in Figure 5.
\begin{figure}[htbp]
\centerline{\includegraphics[width=0.8\textwidth]{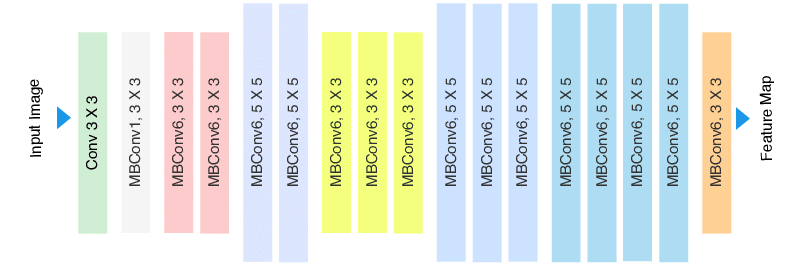}}
\caption{Architechture of EfficientNet \cite{ahmed2022classification}}
\label{fig5}
\end{figure}

EfficientNet B4 was chosen as the base configuration with an input size of 380×380 pixels to balance memory efficiency and performance. Experiments with different variants indicated that EfficientNet B5 provided the best trade-off between accuracy and computational complexity. EfficientNet's ability to capture subtle facial artifacts made it particularly effective for deepfake detection. While smaller models like EfficientNet-B4 offer faster inference times and lower memory consumption, B5 demonstrated improved accuracy in detecting subtle deepfake artifacts, which is critical in this application. Larger variants such as EfficientNet-B6 and B7 were tested but proved impractical due to their significantly higher computational requirements and increased latency during inference. EfficientNet-B5 provided an optimal balance, achieving superior performance while maintaining reasonable computational costs, making it suitable for real-time or near-real-time deepfake detection systems.

\subsection{Prediction Averaging Strategy}
The final prediction for a video was obtained by aggregating frame-level predictions. Each video was represented by predictions from 32 sampled frames. Rather than using simple averaging, a confidence-weighted aggregation approach was employed. Frame-level predictions were assigned different weights based on their confidence scores. Frames with prediction probabilities above a high-confidence threshold were given greater weight, while those below a low-confidence threshold were disregarded. This strategy ensured that the final video-level prediction was more robust, as it prioritized reliable frame-level inferences while minimizing the influence of uncertain predictions.

The combination of robust face detection (MTCNN), efficient feature extraction (EfficientNet), and a confidence-based prediction averaging strategy proved effective in identifying deepfake videos.

\subsection{Training}

The training process was designed to optimize the model’s capability to detect deepfake videos while ensuring computational efficiency. Balanced sampling was employed to maintain an equal number of real and fake face crops in each training batch. Stochastic Gradient Descent (SGD) with momentum was used as the optimizer. A polynomial learning rate scheduler was applied. Batch sizes were adapted based on the EfficientNet variant used. 

To enhance the model’s contextual understanding, a 30\% margin was added around detected face crops during training. This approach allowed the model to capture artifacts in the surrounding facial regions.  Label smoothing was applied to reduce overconfidence in predictions and enhance robustness. 

The validation approach evolved throughout the development process. Initially, folders 0 through 2 were held out for validation. Validation performance was monitored using two log loss metrics: Log loss on real videos and Log loss on fake videos. From the validation, it was found that logloss on FAKE videos was much lower than on REAL.

\section{Result and Discussion}
\subsection{Performance Evaluation Metrics}
\begin{enumerate}
    \item Log loss: Log loss, also known as binary cross-entropy, is a loss function utilized in ML to measure the difference between predicted binary outcomes and actual binary labels. The formula for the log loss function is shown in Equation 1, where \( n \) is the number of predicted videos, \( \hat{y}_i \) is the estimated likelihood that the video is a fake., \( y_i \) is 1 if the video is FAKE, 0 if REAL, and \( \log() \) is the natural logarithm.
    \begin{equation}
        \text{LogLoss} = -\frac{1}{n} \sum_{i=1}^{n} \left[ y_i \log(\hat{y}_i) + (1 - y_i) \log(1 - \hat{y}_i) \right]
    \end{equation}  
    \item F1-score: The F1 score is a performance metric used to evaluate the accuracy of a classification model. It is the harmonic mean of precision and recall as shown in Equation 2.
    \begin{equation}
    F-score=\frac{2*Precision*Recall}{Precision + Recall} 
    \end{equation}
    \item AUC: The performance of the classifier over all potential classification thresholds is summed up by a single figure called the ROC AUC score. The classifier's ability to discern between positive and negative classes is calculated by the score. 
\end{enumerate}

\subsection{Result}
The results obtained using the proposed EfficientNet-B5 model in conjunction with the MTCNN face detector demonstrate competitive performance in deepfake video detection. Specifically, the model achieved a log loss of 0.4278, an AUC of 0.938, and an F1 score of 0.8682, denoting its strong capability to distinguish between real and fake videos with high precision and recall.

In comparison to the work of Coccomini et al. \cite{coccomini2022combining}, who integrated EfficientNet with Vision Transformers, the AUC increased marginally to 0.951, and the F1 score improved to 0.88, highlighting the potential advantage of combining Vision Transformers with CNNs to enhance feature representation and classification performance.

Conversely, the ensemble CNN model proposed by Bonettini et al. \cite{bonettini2021video} yielded a slightly higher log loss of 0.4640. This suggests that, while ensemble models can enhance generalization, their performance may be less robust compared to hybrid approaches.

Overall, these findings underscore the effectiveness of the proposed EfficientNet-B5 model and contribute important perceptions into the comparative benefits of hybrid architectures in addressing the complexities of deepfake video detection. The comparison of results is shown in Table 1.

 \begin{table}[htbp]
\caption{Comparison of Proposed Model with respect to Evaluation Metrics}
\begin{center}
\begin{tabular}{|c|c|c|c|}
\hline
 Model & Log loss & AUC & F1 score \\
 \hline
 Proposed MTCNN-EfficientNetB5 & 0.4278 & 0.9380 & 0.8682 \\
 \hline
 EfficientNet-Vision Transformer \cite{coccomini2022combining} & - & 0.951 & 0.88  \\
 \hline
 Ensemble CNN \cite{bonettini2021video} & 0.464 & - & - \\
 \hline
\end{tabular}
\end{center}
\end{table} 

\subsection{Discussion}
One of the primary advantages of the proposed model is its capacity to attain high classification performance with a relatively lower log loss compared to ensemble models. This suggests that the EfficientNet-B5 architecture, when combined with MTCNN for face detection, is well-suited for the task of deepfake detection due to its efficient feature extraction and the robustness of its facial region localization.

However, the approach is not without its limitations. First, the model's performance, although strong, still falls slightly short of the hybrid EfficientNet-Vision Transformer approach. This suggests that convolution-based architectures may benefit from attention mechanisms to further improve feature extraction. Additionally, the reliance on face detection introduces potential vulnerabilities, such as inaccuracies in face localization or challenges in detecting faces under occlusion or poor lighting conditions. These limitations may hinder the model's performance in more diverse and complex real-world scenarios.

Future work can explore the integration of Transformer or other attention-based models with the EfficientNet-B5 backbone to enhance its feature extraction capabilities. Moreover, developing robust face detection pipelines that can handle occlusions, low-resolution inputs, and variations in lighting conditions would further improve the detection accuracy. Finally, evaluating the model on a broader range of datasets representing diverse manipulation techniques and real-world variations will be essential to ensure its applicability in practical detection systems.

\section{Conclusion}
Deepfake technology has rapidly gained traction with the proliferation of digital content, particularly on social media platforms. As these technologies become increasingly accessible, they pose substantial threats to media authenticity. In response, deep learning has emerged as a robust approach for detecting manipulated content. This paper contributes to deepfake detection by integrating the EfficientNet-B5 model with the MTCNN face detector. This hybrid approach enhances facial region localization and feature extraction, improving detection performance. The primary innovation lies in combining a state-of-the-art CNN with precise face detection to strengthen facial manipulation identification. The results underscore the importance of accurate face region extraction and highlight hybrid methodologies, such as combining CNNs with Vision Transformers, as promising future directions. This work offers insights into architectural choices that can advance more resilient and generalizable deepfake detection systems.

%
%
%
\bibliographystyle{plain}
\bibliography{main.bib}

@article{durall2019unmasking,
  title={Unmasking deepfakes with simple features},
  author={Durall, Ricard and Keuper, Margret and Pfreundt, Franz-Josef and Keuper, Janis},
  journal={arXiv preprint arXiv:1911.00686},
  year={2019}
}

@article{kingra2023emergence,
  title={Emergence of deepfakes and video tampering detection approaches: A survey},
  author={Kingra, Staffy and Aggarwal, Naveen and Kaur, Nirmal},
  journal={Multimedia Tools and Applications},
  volume={82},
  number={7},
  pages={10165--10209},
  year={2023},
  publisher={Springer}
}

@article{abbas2024unmasking,
  title={Unmasking deepfakes: A systematic review of deepfake detection and generation techniques using artificial intelligence},
  author={Abbas, Fakhar and Taeihagh, Araz},
  journal={Expert Systems With Applications},
  pages={124260},
  year={2024},
  publisher={Elsevier}
}

@article{arshed2023unmasking,
  title={Unmasking Deception: Empowering Deepfake Detection with Vision Transformer Network},
  author={Arshed, Muhammad Asad and Alwadain, Ayed and Faizan Ali, Rao and Mumtaz, Shahzad and Ibrahim, Muhammad and Muneer, Amgad},
  journal={Mathematics},
  volume={11},
  number={17},
  pages={3710},
  year={2023},
  publisher={MDPI}
}

@inproceedings{das2023unmasking,
  title={Unmasking Deepfakes: Masked Autoencoding Spatiotemporal Transformers for Enhanced Video Forgery Detection},
  author={Das, Sayantan and Kolahdouzi, Mojtaba and {\"O}zparlak, Levent and Hickie, Will and Etemad, Ali},
  booktitle={2023 IEEE International Joint Conference on Biometrics (IJCB)},
  pages={1--11},
  year={2023},
  organization={IEEE}
}

@article{rana2022deepfake,
  title={Deepfake detection: A systematic literature review},
  author={Rana, Md Shohel and Nobi, Mohammad Nur and Murali, Beddhu and Sung, Andrew H},
  journal={IEEE access},
  volume={10},
  pages={25494--25513},
  year={2022},
  publisher={IEEE}
}

@article{taeb2022comparison,
  title={Comparison of deepfake detection techniques through deep learning},
  author={Taeb, Maryam and Chi, Hongmei},
  journal={Journal of Cybersecurity and Privacy},
  volume={2},
  number={1},
  pages={89--106},
  year={2022},
  publisher={MDPI}
}

@article{janutenas2023deep,
  title={Deep Learning Methods to Detect Image Falsification},
  author={Janut{\.e}nas, Laimonas and Janut{\.e}nait{\.e}-Bogdanien{\.e}, J{\=u}rat{\.e} and {\v{S}}e{\v{s}}ok, Dmitrij},
  journal={Applied Sciences},
  volume={13},
  number={13},
  pages={7694},
  year={2023},
  publisher={MDPI}
}

@article{chu2022protecting,
  title={Protecting world leader using facial speaking pattern against deepfakes},
  author={Chu, Beilin and You, Weike and Yang, Zhen and Zhou, Linna and Wang, Renying},
  journal={IEEE Signal Processing Letters},
  volume={29},
  pages={2078--2082},
  year={2022},
  publisher={IEEE}
}

@article{bansal2023real,
  title={Real-time advanced computational intelligence for deep fake video detection},
  author={Bansal, Nency and Aljrees, Turki and Yadav, Dhirendra Prasad and Singh, Kamred Udham and Kumar, Ankit and Verma, Gyanendra Kumar and Singh, Teekam},
  journal={Applied Sciences},
  volume={13},
  number={5},
  pages={3095},
  year={2023},
  publisher={MDPI}
}

@article{suratkar2023deep,
  title={Deep fake video detection using transfer learning approach},
  author={Suratkar, Shraddha and Kazi, Faruk},
  journal={Arabian Journal for Science and Engineering},
  volume={48},
  number={8},
  pages={9727--9737},
  year={2023},
  publisher={Springer}
}

@inproceedings{agarwal2020detecting,
  title={Detecting deep-fake videos from appearance and behavior},
  author={Agarwal, Shruti and Farid, Hany and El-Gaaly, Tarek and Lim, Ser-Nam},
  booktitle={2020 IEEE international workshop on information forensics and security (WIFS)},
  pages={1--6},
  year={2020},
  organization={IEEE}
}

@article{akhtar2024video,
  title={Video and audio deepfake datasets and open issues in deepfake technology: being ahead of the curve},
  author={Akhtar, Zahid and Pendyala, Thanvi Lahari and Athmakuri, Virinchi Sai},
  journal={Forensic Sciences},
  volume={4},
  number={3},
  pages={289--377},
  year={2024},
  publisher={MDPI}
}

@article{rashid2021blockchain,
  title={Blockchain technology for combating deepfake and protect video/image integrity},
  author={Rashid, Md Mamunur and Lee, Suk-Hwan and Kwon, Ki-Ryong},
  journal={Journal of Multimedia Society},
  volume={24},
  number={8},
  pages={1044--1058},
  year={2021}
}

@article{deng2022deepfake,
  title={Deepfake Video Detection Based on EfficientNet-V2 Network},
  author={Deng, Liwei and Suo, Hongfei and Li, Dongjie},
  journal={Computational Intelligence and Neuroscience},
  volume={2022},
  number={1},
  pages={3441549},
  year={2022},
  publisher={Wiley Online Library}
}

@article{tan2019efficientnet,
  title={Efficientnet: Rethinking model scaling for convolutional neural networks},
  author={Tan, Mingxing},
  journal={arXiv preprint arXiv:1905.11946},
  year={2019}
}

@article{ahmed2022classification,
  title={Classification and understanding of cloud structures via satellite images with EfficientUNet},
  author={Ahmed, Tashin and Sabab, Noor Hossain Nuri},
  journal={SN Computer Science},
  volume={3},
  number={1},
  pages={99},
  year={2022},
  publisher={Springer}
}

@inproceedings{yuan2020minor,
  title={Minor privacy protection through real-time video processing at the edge},
  author={Yuan, Meng and Nikouei, Seyed Yahya and Fitwi, Alem and Chen, Yu and Dong, Yunxi},
  booktitle={2020 29th International Conference on Computer Communications and Networks (ICCCN)},
  pages={1--6},
  year={2020},
  organization={IEEE}
}

@inproceedings{coccomini2022combining,
  title={Combining efficientnet and vision transformers for video deepfake detection},
  author={Coccomini, Davide Alessandro and Messina, Nicola and Gennaro, Claudio and Falchi, Fabrizio},
  booktitle={International conference on image analysis and processing},
  pages={219--229},
  year={2022},
  organization={Springer}
}

@inproceedings{bonettini2021video,
  title={Video face manipulation detection through ensemble of cnns},
  author={Bonettini, Nicolo and Cannas, Edoardo Daniele and Mandelli, Sara and Bondi, Luca and Bestagini, Paolo and Tubaro, Stefano},
  booktitle={2020 25th international conference on pattern recognition (ICPR)},
  pages={5012--5019},
  year={2021},
  organization={IEEE}
}
\end{document}